\documentclass[10pt, a4paper]{article}

\usepackage[final]{lrec2026} 

\usepackage{graphicx}
\usepackage{comment}
\usepackage{longtable}   
\usepackage{array}       
\usepackage{tabularx}    
\usepackage{booktabs}    
\usepackage{courier}     
\usepackage{url}
\usepackage{hyperref}
\usepackage{tikz}
\usepackage{listings}
\usepackage{comment}

\newcommand{\ignore}[1]{}
 \newcounter{todocounter}

\newcounter{todocounterac}

\newcounter{todocounterpe}

\title{Analysing Lightweight Large Language Models for Biomedical Named Entity Recognition on Diverse Ouput Formats}

\name{Pierre Epron$\textsuperscript{1,2}$, Adrien Coulet$\textsuperscript{1*}$\thanks{*These authors share last authorship.}, Mehwish Alam$\textsuperscript{2*}$\footnotemark[1]} 

\address{Inria, Inserm, Université Paris Cité, HeKA, UMR 1346, Paris, France$\textsuperscript{1}$ \\
Télécom Paris, Institut Polytechnique de Paris, France$\textsuperscript{2}$ \\
pierre.epron@inria.fr, adrien.coulet@inria.fr, mehwish.alam@telecom-paris.fr }

\abstract{
Despite their strong linguistic capabilities, Large Language Models (LLMs) are computationally demanding and require substantial resources for fine-tuning, which is unadapted to privacy and budget constraints of many healthcare settings. To address this, we present an experimental analysis focused on Biomedical Named Entity Recognition using lightweight LLMs, 
we evaluate the impact of different output formats on model performance. 
The results reveal that lightweight LLMs can achieve competitive performance compared to the larger models, highlighting their potential as lightweight yet effective alternatives for biomedical information extraction. Our analysis shows that instruction tuning over many distinct formats does not improve performance, but identifies several format consistently associated with better performance.
 \\ \newline \Keywords{Large Language Models, Named Entity Recognition, Biomedicine, Resource-constrained NLP}}

\begin{document}

\maketitleabstract

\section{Introduction}
\label{sec:intro}

Generative Named Entity Recognition (G-NER) offers a promising paradigm shift from traditional span-based or classification-based approaches by framing entity extraction as a text generation task~\cite{xu2024large}. 
In this setting, structured prediction plays a central role as it enables models to generate outputs with predefined structures. For example, in Information Extraction (IE), structured prediction inputs unstructured texts and outputs meaningful structured representations of entities, relations or events.

Recent advances in this field have been made possible by instruction-tuned Large Language Models (LLMs) trained on vast multi-domain datasets~\cite{corr/abs-2304-08085, Zhou00CP24}. While these models enable strong generalization, their large scale is inappropriate for some applications. For example, in the biomedical domain, reducing the scope of the domain could lead to smaller and faster models, better suited to the budget and privacy constraints associated with many applications in healthcare.

In addition to model size, the choice of output format also plays a crucial role. Current research heavily relies on a single output format, such as JSON or a corpus-specific template, yet such fixed formats may introduce bias and restrict usability.

In this work, we investigate instruction tuning on top of lightweight LLM for the task of NER in the biomedical domain, and specifically examine how the size of the LLM and the choice of a single output format influences NER performance.
We designed experiments across twelve different output formats and analyze their impact on model effectiveness. 


Our results indicate that lightweight LLMs are fully capable of performing biomedical NER, and that tuning a format-agnostic, generative system yields only marginal performance gains. This underscores the approach’s flexibility, practical applicability in resource-constrained settings, and inherent robustness to potential biases, making it a compelling alternative to larger, more rigid models. Additionally, we observed that in contrast to our initial intuition, instruction tuning over many distinct formats does not improve performance. However, we identify output format consistently associated with better performance.
This article is organized as follows.

Section~\ref{sec:related-work} discusses related works and positions our contribution w.r.t. the State-of-The-Art (SoTA). Section~\ref{sec:methodology} presents the proposed method, while Section~\ref{sec:experimentation} and~\ref{sec:results} detail experimental analysis for investigating the impact on G-NER of model size and output format. Finally, sections~\ref{sec:conclusion} and ~\ref{sec:limitations} conclude and discuss limitations of the current work. You can find the code on this repository: \href{https://github.com/PierreEpron/MF-NER}{https://github.com/PierreEpron/MF-NER}

\section{Related Work}
\label{sec:related-work}

Since the rise of LLMs there has been remarkable progress in the task of IE. This section discusses most recent methods for NER based on LLMs. Others are excluded for the consistency of  comparison and space limitation. See Xu et al.~\cite{XuCPZXZWZWC24} for a recent and detailed survey. 

\paragraph{Prompting based Approaches.}
Zero-shot Information Extraction (ZIE) aims to reduce reliance on annotated data while maintaining strong performance. 
Inspired by LLMs such as GPT-3 and ChatGPT, ChatIE~\cite{corr/abs-2302-10205} reformulates ZIE as a multi-turn question-answering problem using a two-stage framework. It is evaluated on NER, entity-relation extraction and event extraction tasks across multiple languages. ChatIE achieves performance competitive with fully supervised models, demonstrating the potential of resource-efficient (i.e., without training) IE approaches based only on inferencing. 
In parallel, LLMs with code-style prompts have demonstrated notable success. In ~\cite{talip/BiCJXGCZ24}, the authors frame knowledge graph construction as a code completion task, utilizing schema-aware prompts and rationale-enhanced generation to improve extraction accuracy. 

Few-shot learning approaches address cross-domain adaptation with limited data. GPT-NER~\cite{wang2023gpt} recognizes entities by surrounding them with the special characters \texttt{\#} and \texttt{@}. They evaluate multiple solutions of few-shot demonstration retriever and report their best results which were achieved using entity level embeddings. Paolini et al.~\cite{PaoliniAKMAASXS21} unify structured generation for diverse IE tasks, while Chen et al.~\cite{ChenLQ0TJHC23} propose a collaborative domain-prefix tuning for cross-domain NER. 

CodeIE~\cite{LiSTYWHQ23} leverages code-style prompts with in-context examples to achieve superior few-shot performance. Inverse generation, where structured data is converted into text or questions, has emerged as another effective strategy. SynthIE~\cite{JosifoskiSP023} generates high-quality synthetic data by reversing task directions, enabling downstream models to surpass previous benchmarks on the task of relation extraction.

\paragraph{RAG based Approaches.}
Retrieval Augmented Generation (RAG) based methods enhance model performance by leveraging auxiliary knowledge. Li et al.~\cite{LiLPSWZP23} introduce a two-stage multimodal NER framework that heuristically retrieves refined knowledge to improve entity prediction, while Amalvy et al.~\cite{AmalvyLD23} generate synthetic context datasets and train neural retrievers to support NER on long documents. 
Code4UIE~\cite{GuoLJLZLLYBGC24} addresses the challenges of non-unified prompts and limited in-context learning by introducing a Python class–based schema that standardizes diverse IE tasks and incorporates retrieval-augmented mechanisms. 



\paragraph{Instruction Tuning based Approaches.}
 \cite{HuJLHXHWY23} propose entity-to-text augmentation by manipulating entity lists and employing diversity beam search to improve dataset richness for NER. Instruction tuning benefits from these synthetic datasets as well. UniNER~\cite{Zhou00CP24} distills ChatGPT into smaller student models through mission-focused instruction tuning for open NER, while Ding et al.~\cite{DingLWTBZ24} incorporates negative instances in generative NER training on \emph{The Pile} open-source corpus, improving zero-shot performance on unseen entity domains.
Supervised fine-tuning approaches further enhance model capabilities. DeepStruct~\cite{WangLCH0S22} pretrains LLMs on task-agnostic corpora to improve structural understanding, while GIELLM~\cite{corr/abs-2311-06838} fine-tunes LLMs on mixed datasets for Japanese IE, leveraging mutual reinforcement effects to improve performance across multiple tasks. GoLLIE~\cite{iclr/SainzGALRA24} fine-tunes LLMs with the instructions related to the annotation guidelines across a small set of IR tasks. 

Wang et al.~\cite{corr/abs-2304-08085} model IE as instruction-guided text generation. An option mechanism and auxiliary tasks refine span, relation, and event extraction. This improves structural and semantic understanding.



\paragraph{Other Approaches.}
Iterative self-improvement approaches such as ProgGen~\cite{HengDLYLZZ24} guide LLMs through self-reflection to generate domain-relevant attributes and proactively construct NER context data, improving the quality of generated datasets.
Constrained decoding strategies have been used to ensure structured outputs, e.g., ~\cite{GengJP023} introduce grammar-constrained decoding with input-dependent grammars, while~\cite{ZaratianaTHC24} propose a text-to-graph framework generating linearized graphs with a transformer encoder-decoder architecture, a pointing mechanism and dynamic vocabularies for joint entity and relation extraction.

While existing studies have explored various methods for G-NER, our work experimentally demonstrates that lightweight LLMs, when used with suitable output formats, can achieve performance competitive with larger models.


\section{Methodology}
\label{sec:methodology}


In this work, we focus on Causal Language Models (CLMs), which generate text autoregressively by predicting each token given its preceding context. Trained with the Next Token Prediction (NTP) objective, CLMs naturally align with text generation tasks, making them well-suited for instruction tuning. We adopt this framework to treat NER as a text generation problem, where entities are produced directly as model outputs. Compared to encoder-decoder architectures like BART, CLMs are more compatible with instruction-following objectives.


\subsection{Instruction Tuning}


Instruction tuning involves fine-tuning a model on a dataset specifically designed for a given downstream task, in this case, NER. Such datasets comprise pairs composed of one natural language instruction and its corresponding expected outputs. Formally, these datasets are represented as:

\begin{equation}
\label{equation:eq1}
    D_{instr} = \{(x_i,y_i)\}_{i=1}^{N}
\end{equation}

where $x_i$ denotes a natural language instruction, $y_i$ is the corresponding target output, and $N$ is the total number of instruction–response pairs. Instruction-tuning can further be adapted for the task of NER as follows~\cite{Zhou00CP24}: 

\begin{equation}
\label{equation:eq2}
D_{instr} = \{(x_i, y_i, t_i, d_i)\}_{i=1}^{N}    
\end{equation}

where $t_i$ represents the $i-th$ entity type to be recognized by the model and $d_i$ represents a document used to encode characteristics specific to the dataset. 
For example, this document be an annotation guidelines that specifies recommendations for ambiguous cases such as ``lung tumor'' that can either be considered as an \texttt{BODY PART} or as a \texttt{DISEASE} depending on the context.
Similarly, we add $f_i$ that represents the specified output format to enrich the instruction-tuning dataset as follow:


 \begin{equation}
 \label{equation:eq3}
 D_{instr} = \{(x_i, y_i, t_i, d_i, f_i)\}_{i=1}^{N}  
 \end{equation}

\subsection{Formats}

In this study, a format refers to the specific manner in which retrieved entities are represented within the system output. Each format defines a unique encoding structure. For example, the BIO (Beginning, Inside, Outside) is a widely adopted format for representing entities in annotated corpora used for training and evaluating supervised NER models. For this study, we selected twelve formats that ensure broad coverage of commonly used representation variants. We tested these different formats in different configurations of instruction tuning: training with either a single format or by mixing several (see Section \ref{sec:training_config} for the configurations). The rationale behind our choices is detailed below, with an example corresponding to the user prompt as shown in Listings \ref{lst:prompt1} and \ref{lst:prompt1example}.

\begin{lstlisting}[breaklines=true, caption={Prompt}, label={lst:prompt1}, breakindent=0pt, basicstyle=\ttfamily, showstringspaces=false, frame=single]
The task you need to complete is named entity recognition. Follow {dataset} guidelines.

{format}

Text: {text}
\end{lstlisting}

\begin{lstlisting}[breaklines=true, breakindent=0pt, caption={Prompt Arguments}, label={lst:prompt1example}, basicstyle=\ttfamily, showstringspaces=false, frame=single]
dataset = Genia
text = These results suggest that BCL6 plays a role in activated lymphocytes as an immediate early gene.
\end{lstlisting}






The following \emph{conv\_term} and \emph{single\_tag} formats are selected because those are used in the SoTA approaches, i.e., UniNER and GPT-NER respectively.
We also selected a \emph{multi\_tag} format that uses standard XML tags (e.g., <person></person>).


\vspace{11pt}

\noindent\textbf{conv\_term}

\begin{lstlisting}[breaklines=true, breakindent=0pt, basicstyle=\ttfamily, showstringspaces=false]
user :: Type: cell_line
assistant :: Answer: []
user :: Type: protein
assistant :: Answer: ["BCL6"]
\end{lstlisting}

\noindent\textbf{single\_tag}

\begin{lstlisting}[breaklines=true, breakindent=0pt, basicstyle=\ttfamily, showstringspaces=false]
assistant :: Answer: These results suggest that BCL6 plays a role in @@activated @@lymphocytes\#\#\#\# as an immediate early gene.
\end{lstlisting}

\noindent\textbf{multi\_tag} 

\begin{lstlisting}[breaklines=true, breakindent=0pt, basicstyle=\ttfamily, showstringspaces=false]
assistant :: Answer: These results suggest that <protein>BCL6</protein> plays a role in <cell_type>activated <cell_type>lymphocytes</cell_type></cell_type> as an immediate early gene.
\end{lstlisting}

The \emph{single\_code} and \emph{multi\_code} formats are selected because they outperform other formats in code completion tasks~\cite{li2023codeie}. Although our work does not directly address code completion, we include these formats for completeness and fair comparison, as they are supported by recent LLMs.

\vspace{11pt}

\noindent\textbf{single\_code}

\begin{lstlisting}[breaklines=true, breakindent=0pt, basicstyle=\ttfamily, showstringspaces=false]
assistant :: Answer: These results suggest that <protein>BCL6</protein> plays a role in <cell_type>activated <cell_type>lymphocytes</cell_type></cell_type> as an immediate early gene.
\end{lstlisting}

\noindent\textbf{multi\_code}

\begin{lstlisting}[breaklines=true, breakindent=0pt, basicstyle=\ttfamily, showstringspaces=false]
assistant :: Answer:
```py
def named_entity_recognition(input_text): 
        """ extract entities from the input_text. """
        input_text = "These results suggest that BCL6 plays a role in activated lymphocytes as an immediate early gene."
        entity_list = [] 
        # extracted entities for cell_line, protein, RNA, DNA, cell_type types.
        entity_list.append({"text": "activated lymphocytes", "type": "cell_type"})
        entity_list.append({"text": "BCL6", "type": "protein"})
        entity_list.append({"text": "lymphocytes", "type": "cell_type"})
```
\end{lstlisting}


We select \emph{single\_term} and \emph{multi\_term} because they are standard formats for string extraction. Similarly, for the segment extraction task, we choose \emph{single\_span} and \emph{multi\_span}. Since triple extraction is a common task within IE, we add \emph{multi\_triple} format. Indeed, NER can be seen as a triple extraction where the predicate of the triple is forced to be ``rdf:type" or ``is a". 

\vspace{11pt}

\noindent\textbf{single\_term}

\begin{lstlisting}[breaklines=true, breakindent=0pt, basicstyle=\ttfamily, showstringspaces=false]
assistant :: Answer: ["activated lymphocytes", "lymphocytes"]
\end{lstlisting}

\noindent\textbf{multi\_term}

\begin{lstlisting}[breaklines=true, breakindent=0pt, basicstyle=\ttfamily, showstringspaces=false]
assistant :: Answer: [{"text": "activated lymphocytes", "type": "cell_type"}, {"text": "BCL6", "type": "protein"}, {"text": "lymphocytes", "type": "cell_type"}]
\end{lstlisting}

\noindent\textbf{single\_span}

\begin{lstlisting}[breaklines=true, breakindent=0pt, basicstyle=\ttfamily, showstringspaces=false]
assistant :: Answer: [[48,69],[58,69]]} 
\end{lstlisting}

\noindent\textbf{multi\_span}

\begin{lstlisting}[breaklines=true, breakindent=0pt, basicstyle=\ttfamily, showstringspaces=false]
assistant :: Answer: [{"span": [48, 69], "type": "cell_type"}, {"span": [27, 31], "type": "protein"}, {"span": [58, 69], "type": "cell_type"}]
\end{lstlisting}

\noindent\textbf{multi\_triple}

\begin{lstlisting}[breaklines=true, breakindent=0pt, basicstyle=\ttfamily, showstringspaces=false]
assistant :: Answer:
activated lymphocytes; is a; cell_type
BCL6; is a; protein
lymphocytes; is a; cell_type
\end{lstlisting}

Finally, \emph{multi\_bio} and \emph{multi\_brat} are well known annotation schema that model might have process multiple time during its pretraining or first instruction tuning.

\vspace{11pt}

\noindent\textbf{multi\_bio}

\begin{lstlisting}[breaklines=true, breakindent=0pt, basicstyle=\ttfamily, showstringspaces=false]
assistant :: Layer 1: O O O O B-protein O O O O B-cell_type I-cell_type O O O O O
Layer 2: O O O O O O O O O O B-cell_type O O O O O\end{lstlisting}

\noindent\textbf{multi\_brat}

\begin{lstlisting}[breaklines=true, breakindent=0pt, basicstyle=\ttfamily, showstringspaces=false]
assistant :: Answer:
T1        cell_type 48 69        activated lymphocytes
T2        protein 27 31        BCL6
T3        cell_type 58 69        lymphocytes
\end{lstlisting}

Please note that all considered formats enable the representation of nested entities, but that \emph{single\_span}, \emph{multi\_span}, and \emph{multi\_bio} do not enable representing discontinuous entities. \emph{single\_span} cannot be adapted to discontinuous entities, but the two other formats could by relying on indexing and BIOHD. However, we avoided these adaptations for the sake of simplicity. 

\subsection{Training}


\paragraph{Model.}

We parameterize the conditional distribution of the output sequence given the input instruction using an auto-regressive language model:

\begin{equation}
    P_{\theta}(y|x) = \prod_{t=1}^T P_{\theta}(y_t|x,y_{< t})
\end{equation}

where $\theta$ denotes the learnable parameters of the model, $T$ is the length of the output sequence $y$, $y_t$ is the token at position $t$, and $y_{<t}$ represents the preceding tokens in the sequence. 

\paragraph{Loss Function.}

The training objective is based on minimizing the negative log-likelihood of the correct output sequence given the instruction. For a single instruction–response pair $(x,y)$, the instruction-tuning loss is defined as:

\begin{equation}
    \mathcal{L}_{instr}(x,y) = - \sum_{t=1}^T log P_{\theta} (y_t | x,y_{<t})
\end{equation}

which corresponds to the standard cross-entropy loss over tokens. More explicitly, the cross-entropy formulation is given by:

\begin{equation}
    \mathcal{L}_{CE} = - \frac{1}{T} \sum_{t=1}^T \sum_{i=1}^\mathcal{V} y_i^{(t)} log~\hat{y}_{i}^{(t)}
\end{equation}

where $\mathcal{V}$ is the vocabulary size, $y_i^{(t)}$ is the one-hot encoding of the correct token at step $t$, and $\hat{y}_i^{(t)}$ is the vector of predicted probabilities distribution over the vocabulary.









\section{Experimental Setting}
\label{sec:experimentation}


\subsection{Datasets}

We selected eight BioNER datasets for our analysis. 
See Table~\ref{tab:datasets} for detailed statistics of each dataset.

\noindent \textbf{AnatEM}~\cite{pyysalo2014anatomical} 
contains PubMed abstracts, annotated for anatomical entities, including organs, tissues, and body parts.

\noindent \textbf{BioCreative II Gene Mention (BC2GM)}~\cite{smith2008overview} is a PubMed-based corpus annotated for gene and protein mentions. 


\noindent \textbf{BC4CHEMD}~\cite{krallinger2015chemdner} is also a PubMed-based corpus annotated for chemical compound and drug names.


\noindent \textbf{BC5CDR}~\cite{li2016biocreative} is a benchmark corpus for BioNER and relation extraction, focusing on chemical and disease entities. 

\noindent \textbf{CADEC}~\cite{karimi2015cadec} is a corpus of user-generated generated content from forum posts about medication experience (AskAPatient.com) annotated for drugs, adverse events, and related attributes.


\noindent \textbf{GENIA}~\cite{kim2003genia} is a MEDLINE-derived corpus manually annotated for proteins, DNA, RNA, cell types, and cell lines using a structured ontology.


\noindent \textbf{NCBI Disease}~\cite{dougan2014ncbi} is a PubMed-based corpus manually annotated for disease mentions, normalized to MEDIC identifiers.


\noindent \textbf{PGxCorpus}~\cite{legrand2020pgxcorpus} is a PubMed-based pharmacogenomics corpus annotated for genes, drugs, phenotypes, and their interactions.


\begin{table*}[!h]
\begin{center}
\begin{tabular}{@{}llllccl@{}}
\toprule
dataset & train & dev & test & nested & discont & labels \\ \midrule
AnatEM & 5861 & 2118 & 3830 & - & - & 1 \\
bc2gm & 12500 & 2500 & 5000 & - & - & 1 \\
bc4chemd & 30682 & 30639 & 26364 & - & - & 1 \\
bc5cdr & 4560 & 4581 & 4797 & - & - & 2 \\
CADEC & 5317 & 1140 & 1140 & - & $\sim$6.5-8.5\% & 5 \\
GENIA & 15023 & 1669 & 1854 & $\sim$2-2.5\% & - & 5 \\
ncbi & 5432 & 923 & 940 & - & - & 1 \\
PGx & 661 & 142 & 142 & $\sim$4-4.5\% & $\sim$0.8-1.8\% & 10 \\ \bottomrule
\end{tabular}
\caption{Distribution of datasets used for this work. Columns nested and discount are a range approximation. Example for GENIA\_NER, each split have between 2 and 2.5 percent of nested entities.}
\label{tab:datasets}
\end{center}
\end{table*}



\subsection{Baselines}
\label{sec:baselines}

We compare our performances against UniNER and InstructIE (See Section~\ref{sec:related-work}). The baselines are further complemented with a standard BERT model and GLiNER, when available. GLiNER is a lightweight, open-source NER model that leverages a bidirectional transformer encoder, such as BERT. Unlike conventional NER systems restricted to a fixed set of entity types, GLiNER supports a dynamic set of entity types using natural language prompts, making it well suited for zero-shot generalization across diverse domains. For PGxCorpus the baseline reported is from the original study, which employed Convolutional Neural Network for evaluation. For CADEC, we compare against the state-of-the-art Grid-Tagging approach~\cite{liu2022toe}, which frames the task as a classification of relationships between words as head, tail, or neighbor.


\subsection{Language Models}


We first conducted experiments using Qwen2.5-0.5B-Instruct\footnote{\href{https://huggingface.co/Qwen/Qwen2.5-0.5B-Instruct}{Qwen2.5-0.5B-Instruct}}
, a relatively lightweight LLM that demonstrates strong performance \cite{qwen2}. To enable a reliable comparison between formats, it was important to select a model that could be trained multiple times without excessive computational cost. Following the same rationale, we replicated our experiments with Llama-3.2-1B-Instruct\footnote{\href{https://huggingface.co/meta-llama/Llama-3.2-1B-Instruct}{Llama-3.2-1B-Instruct}}, a more recent model from the Meta LLaMA family \cite{dubey2024llama} featuring twice the number of parameters. We consider these two models as lightweight with regards to larger ones used in baseline experiments such as \cite{touvron2023llama, chiang2023vicuna, chung2024scaling}.

\subsection{Training Configuration}
\label{sec:training_config}

We perform instruction tuning experiments using different format con-
figurations. In the following we describe the formats
used during training:
    \begin{itemize}
        \item \emph{all} includes all the formats defined above in a balanced way.
        \item \emph{7best} mixes only the seven best formats from the best results obtained with the \emph{all} configuration (conv\_term, multi\_tag, multi\_term, multi\_triple, single\_code, single\_tag, single\_term);
        \item \emph{term\_ner} includes \emph{multi\_term} and \emph{single\_term} and was used to evaluate how the use of the same multi and single formats could improve each other's performance when used in combination;
        \item Finally, the \emph{only} configuration concerns models trained with a single format only. 
        We did this training for the two best formats: \emph{conv\_term}, \emph{multi\_triple}. And with  \emph{multi\_term} and \emph{single\_term} to compare with \emph{term\_ner}.
        
    \end{itemize}

    Following the experimental settings in UniNER and InstructIE, each split is a sampling since the total number of examples is too large. 
    For the training set, we took a maximum of $10,000$ examples per dataset. We sampled $200$ examples for the development set and $300$ examples for the test set. For train and dev, we randomly selected one format depending on the configuration of the format and the compatibility of the dataset. We constrained examples from datasets containing discontinuous entities not to be associated with incompatible formats (\emph{multi\_bio}, \emph{multi\_tag}, \emph{single\_tag}). To ensure a robust evaluation, we tested each format from each configuration on the full test set.

    The training hyper-parameters were chosen similar to UniNER and InstructIE. 
    We trained each model for 4 epochs with a validation for each 1/4 of epoch. The validation is evaluated with micro-f1 and we used the best checkpoint for testing. 

    We used micro precision, recall, and f1 for both validation and test as those are standard metrics for NER. In order to evaluate the variability of the sampling procedure, we draw 3 samples for train, dev, and test. Each format configuration is trained and tested on each of the 3 samples and we reported the mean and standard deviation for each metrics over the 3 draws. 
    
 
\section{Results}
\label{sec:results}




The experimental results obtained indicate two primary findings. First, it is observed that despite the discrepancy in model size between the two base models (500M for Qwen-2.5 and 1B for Llama-3.2), the performance does not significantly decline. This is supported by the fact that the same applies to the baselines considered, which are all based on the models with 7B+ number of parameters. In the medical field, lightweight LLMs appear to be adequate.
Secondly, the maximum difference between a format trained alone and trained jointly other formats is 0.05, which is residual. As a consequence, training with several formats simultaneously does not compromise their performance. This creates flexible models for different real-world applications.

\subsection{Overall results}

\begin{table*}[!h]
\begin{center}
\begin{tabular}{{@{{}}lccc|ccc@{{}}}}
\toprule
x & \multicolumn{3}{c}{Qwen2.5-0.5B} & \multicolumn{3}{c}{Llama-3.2-1B} \\
\midrule
 & P$\uparrow$ & R$\uparrow$ & F1$\uparrow$ & P$\uparrow$ & R$\uparrow$ & F1$\uparrow$ \\
\midrule
all & 0.65  \scriptsize ($\pm$0.01) & 0.52  \scriptsize ($\pm$0.01) & 0.58  \scriptsize ($\pm$0.01) & 0.67  \scriptsize ($\pm$0.01) & 0.54  \scriptsize ($\pm$0.01) & 0.60  \scriptsize ($\pm$0.01)\\
7best & 0.81  \scriptsize ($\pm$0.01) & 0.75  \scriptsize ($\pm$0.01) & 0.78  \scriptsize ($\pm$0.01) & \underline{\textbf{0.82}}  \scriptsize ($\pm$0.00) & 0.74  \scriptsize ($\pm$0.01) & 0.78  \scriptsize ($\pm$0.00)\\
multi\_triple & 0.79  \scriptsize ($\pm$0.02) & 0.68  \scriptsize ($\pm$0.01) & 0.73  \scriptsize ($\pm$0.01) & 0.80  \scriptsize ($\pm$0.01) & 0.76  \scriptsize ($\pm$0.01) & 0.78  \scriptsize ($\pm$0.01)\\
conv\_term & 0.81  \scriptsize ($\pm$0.01) & \underline{\textbf{0.79}}  \scriptsize ($\pm$0.01) & \underline{\textbf{0.80}}  \scriptsize ($\pm$0.00) & 0.81  \scriptsize ($\pm$0.01) & \underline{\textbf{0.79}}  \scriptsize ($\pm$0.01) & \underline{\textbf{0.80}}  \scriptsize ($\pm$0.01)\\
term & 0.81  \scriptsize ($\pm$0.01) & 0.71  \scriptsize ($\pm$0.02) & 0.76  \scriptsize ($\pm$0.01) & 0.81  \scriptsize ($\pm$0.01) & 0.73  \scriptsize ($\pm$0.01) & 0.77  \scriptsize ($\pm$0.01)\\
multi\_term & \underline{\textbf{0.82}}  \scriptsize ($\pm$0.01) & 0.70  \scriptsize ($\pm$0.02) & 0.75  \scriptsize ($\pm$0.01) & 0.82  \scriptsize ($\pm$0.01) & 0.70  \scriptsize ($\pm$0.02) & 0.75  \scriptsize ($\pm$0.01)\\
single\_term & 0.78  \scriptsize ($\pm$0.01) & 0.76  \scriptsize ($\pm$0.02) & 0.77  \scriptsize ($\pm$0.01) & 0.80  \scriptsize ($\pm$0.02) & 0.75  \scriptsize ($\pm$0.01) & 0.77  \scriptsize ($\pm$0.01)\\
\bottomrule
\end{tabular}
\caption{Precision (P), recall (R), f-score (F1) for each model and formats configuration. \textbf{Bold} is for best for each model. \underline{Underline} is for best overall.}
\label{tab:overall_results}
\end{center}
\end{table*}

\begin{table*}[!h]
\begin{center}
\begin{tabular}{{@{{}}lccc|ccc@{{}}}}
\toprule
 & \multicolumn{3}{c}{Qwen2.5-0.5B} & \multicolumn{3}{c}{Llama-3.2-1B} \\
\midrule
 & all$\uparrow$ & 7best$\uparrow$ & only$\uparrow$ & all$\uparrow$ & 7best$\uparrow$ & only$\uparrow$ \\
\midrule
conv\_term & 0.78  \scriptsize ($\pm$0.00) & 0.78  \scriptsize ($\pm$0.01) & \underline{\textbf{0.80  \scriptsize ($\pm$0.00)}} & 0.79  \scriptsize ($\pm$0.01) & 0.79  \scriptsize ($\pm$0.01) & \underline{\textbf{0.80  \scriptsize ($\pm$0.01)}}\\
multi\_brat & 0.13  \scriptsize ($\pm$0.01) & - & - & 0.17  \scriptsize ($\pm$0.01) & - & -\\
multi\_code & 0.56  \scriptsize ($\pm$0.01) & - & - & 0.57  \scriptsize ($\pm$0.01) & - & -\\
multi\_span & 0.10  \scriptsize ($\pm$0.01) & - & - & 0.12  \scriptsize ($\pm$0.01) & - & -\\
multi\_term & 0.74  \scriptsize ($\pm$0.01) & \underline{\textbf{0.75  \scriptsize ($\pm$0.01)}} & \underline{\textbf{0.75  \scriptsize ($\pm$0.01)}} & 0.74  \scriptsize ($\pm$0.01) & 0.74  \scriptsize ($\pm$0.01) & \underline{\textbf{0.75  \scriptsize ($\pm$0.01)}}\\
multi\_triple & \textbf{0.78  \scriptsize ($\pm$0.01)} & \textbf{0.78  \textbf{\scriptsize ($\pm$0.01)}} & 0.73  \scriptsize ($\pm$0.01) & \underline{\textbf{0.79  \scriptsize ($\pm$0.01)}} & \underline{\textbf{0.79  \scriptsize ($\pm$0.00)}} & 0.78  \scriptsize ($\pm$0.01)\\
single\_code & \textbf{0.73  \scriptsize ($\pm$0.01)} & \textbf{0.73  \scriptsize ($\pm$0.01)} & - & \underline{\textbf{0.74  \scriptsize ($\pm$0.01)}} & 0.73  \scriptsize ($\pm$0.01) & -\\
single\_span & 0.11  \scriptsize ($\pm$0.01) & - & - & 0.13  \scriptsize ($\pm$0.01) & - & -\\
single\_term & 0.77  \scriptsize ($\pm$0.01) & \textbf{0.78  \scriptsize ($\pm$0.01)} & 0.77  \scriptsize ($\pm$0.01) & \underline{\textbf{0.79  \scriptsize ($\pm$0.00)}} & 0.78  \scriptsize ($\pm$0.00) & 0.77  \scriptsize ($\pm$0.01)\\
\bottomrule
\end{tabular}
\caption{F-score for each base model and formats with respect to their training configuration. \textbf{Bold} is for best for each model. \underline{Underline} is for best overall.}
\label{tab:format_results}
\end{center}
\end{table*}

\begin{table}[!h]
\begin{center}
\begin{tabular}{lc|c}
\toprule
 & Qwen2.5-0.5B & Llama-3.2-1B \\
\midrule
conv\_term & 0.81  \scriptsize ($\pm$0.01) & 0.82  \scriptsize ($\pm$0.01)\\
multi\_bio & 0.44  \scriptsize ($\pm$0.07) & 0.60  \scriptsize ($\pm$0.03)\\
multi\_tag & 0.82  \scriptsize ($\pm$0.01) & \underline{\textbf{0.83  \scriptsize ($\pm$0.01)}}\\
single\_tag & \underline{\textbf{0.83  \scriptsize ($\pm$0.01)}} & \underline{\textbf{0.83  \scriptsize ($\pm$0.01)}}\\
\bottomrule
\end{tabular}
\end{center}
\caption{F-score without discontinuous datasets for each base model and formats with respect to their training configuration. \textbf{Bold} is for best for each model. \underline{Underline} is for best overall.}
\label{tab:nodiscont_format_results}
\end{table}

In Table~\ref{tab:overall_results}, we can see that even though Llama-3.2 can perform slightly better, Qwen-2.5 performance are very similar regardless of the configuration. For the best configuration, the F1 score of \emph{conv\_term} is 0.80 for both models. We also observe that the performances of the configuration \emph{all} are lower due to some low performance of some formats. More specifically, Table~\ref{tab:format_results} shows that the F1 scores of \emph{multi\_brat}, \emph{multi\_span}, and \emph{single\_span} are less than 0.15. The effectiveness of these formats depends on the model’s ability to extract character spans, a task that yielded suboptimal results in our case potentially due to the use of lightweight LLMs. 
Finally, we can see that the performance of each format does not decline when there are trained with multiple format configurations.


Table~\ref{tab:nodiscont_format_results} compares formats that are not compatible with discontinuous entities with the best overall formats. 
We observe good performance for \emph{single\_tag} and \emph{multi\_tag} formats, i.e., 0.83 and 0.82 respectively. While these two formats seem more appropriate for simple entities, they have the disadvantage of not covering all entities.



\subsection{Baseline comparison}

\begin{table*}[!h]
\begin{center}
\begin{tabular}{@{}lccccccc@{}}
\toprule
 & \multicolumn{2}{c}{Ours} & \multicolumn{3}{c}{Supervised} & \multicolumn{2}{c}{ZeroShot} \\ \midrule
 & 7best$\uparrow$  & \multicolumn{1}{c|}{conv\_term$\uparrow$} & Base$\uparrow$  & UIE$\uparrow$  & \multicolumn{1}{c|}{UniNER$\uparrow$ } & UniNER$\uparrow$  & GliNER$\uparrow$  \\ \midrule
AnatEM & 0.87 & \multicolumn{1}{c|}{0.87} & 0.86 & \textbf{0.89} & \multicolumn{1}{c|}{\textbf{0.89}} & 0.25 & 0.34 \\
GENIA & 0.72 & \multicolumn{1}{c|}{0.77} & 0.73 & 0.76 & \multicolumn{1}{c|}{\textbf{0.78}} & 0.54 & 0.56 \\
bc2gm & 0.79 & \multicolumn{1}{c|}{0.80} & 0.81 & 0.81 & \multicolumn{1}{c|}{\textbf{0.82}} & 0.46 & 0.48 \\
bc4chemd & 0.85 & \multicolumn{1}{c|}{0.87} & 0.87 & 0.88 & \multicolumn{1}{c|}{\textbf{0.89}} & 0.48 & 0.43 \\
bc5cdr & 0.86 & \multicolumn{1}{c|}{0.86} & 0.85 & \textbf{0.89} & \multicolumn{1}{c|}{\textbf{0.89}} & 0.68 & 0.66 \\
ncbi & \textbf{0.88} & \multicolumn{1}{c|}{0.87} & 0.80 & 0.86 & \multicolumn{1}{c|}{0.87} & 0.60 & 0.62 \\
CADEC & 0.70 & \multicolumn{1}{c|}{\textbf{0.76}} & 0.74 & - & \multicolumn{1}{c|}{-} & - & - \\
PGX & 0.62 & \multicolumn{1}{c|}{\textbf{0.71}} & 0.59 & - & \multicolumn{1}{c|}{-} & - & - \\ \bottomrule
\end{tabular}
\caption{F-score of our best model vs. the baselines for each dataset. Best results are in \textbf{bold}.}
\label{tab:baseline_results}
\end{center}
\end{table*}

In Table~\ref{tab:baseline_results}, we see that UniNER slightly outperforms all our models on each dataset, despite being trained on multiple domains and in a single format. However, UniNER uses a model with 7 billion parameters, whereas our models have a maximum of 1 billion. Additionally, our models are tuned solely on biomedical domain, and their stability is demonstrated by the low standard deviation with three samples. From these observations, we can conclude that lightweight LLMs perform similarly than larger ones on a specific domain with less computation time.


\subsection{Limitations of \emph{multi\_triple}}


\begin{table*}[!h]
\begin{center}
\begin{tabular}{{@{{}}lccc|ccc@{{}}}}
\toprule
 & \multicolumn{3}{c}{Qwen2.5-0.5B} & \multicolumn{3}{c}{Llama-3.2-1B} \\
\midrule
 & P$\uparrow$ & R$\uparrow$ & F1$\uparrow$ & P$\uparrow$ & R$\uparrow$ & F1$\uparrow$ \\
\midrule
all & 0.88  \scriptsize ($\pm$0.03) & 0.86  \scriptsize ($\pm$0.03) & 0.87  \scriptsize ($\pm$0.03) & 0.87  \scriptsize ($\pm$0.02) & 0.85  \scriptsize ($\pm$0.02) & 0.86  \scriptsize ($\pm$0.01)\\
7best & 0.87  \scriptsize ($\pm$0.02) & 0.87  \scriptsize ($\pm$0.02) & 0.87  \scriptsize ($\pm$0.02) & 0.88  \scriptsize ($\pm$0.01) & 0.86  \scriptsize ($\pm$0.03) & 0.87  \scriptsize ($\pm$0.02)\\
only & \textbf{0.80}  \scriptsize ($\pm$0.06) & \textbf{0.03}  \scriptsize ($\pm$0.03) & \textbf{0.06}  \scriptsize ($\pm$0.06) & 0.84  \scriptsize ($\pm$0.06) & 0.78  \scriptsize ($\pm$0.03) & 0.81  \scriptsize ($\pm$0.04)\\
\bottomrule
\end{tabular}
\caption{Precision (P), Recall (R), F1-score (F1) for each training configuration of the multi\_triple format. \textbf{Bold} are worst results for each configuration.}
\label{tab:triple_results}
\end{center}
\end{table*}


In Table~\ref{tab:baseline_results}, we observe that the format \emph{multi\_triple} has an f-score close to 0 for the AnatEM dataset. In Table \ref{tab:triple_results}, we observe that this is due to the recall of Qwen \emph{only} training configuration. When looking at the results, in majority of the cases the terms were correctly extracted but the assigned entity type was ``disease" when the only possible label of AnatEM is ``anatomy" (see Listing~\ref{lst:dataset-example}).

\begin{lstlisting}[breaklines=true, caption={Example of the AnatEM Dataset}, label={lst:dataset-example}, breakindent=0pt, basicstyle=\ttfamily, showstringspaces=false, frame=single]
Text: The mechanisms that confer this rapid metastatic capacity to lung tumors are unknown .
Entity: (lung tumors, anatomy)
Triple extracted: lung tumors; is a; disease
\end{lstlisting}


In general, it is semantically correct to tag the term as a disease. But the instruction specifically refers to the anatomy type. The model clearly fails to learn how to use the requested type for the \emph{multi\_triple} format. It still performs as intended in \emph{all} and \emph{7best} configurations, leading to the conclusion that the usage of multiple formats help in learning the appropriate type in such specific configuration. 
 
\subsection{Complex Entities}
\label{sec:complex_results}


\begin{table*}[!h]
\begin{center}
\begin{tabular}{@{}llcc|cc@{}}
\toprule
 & & \multicolumn{2}{c}{Qwen2.5-0.5B} & \multicolumn{2}{c}{Llama-3.2-1B} \\
\midrule
 & & $\Delta$ Nested$\downarrow$ & $\Delta$ Discont$\downarrow$ & $\Delta$ Nested$\downarrow$ & $\Delta$ Discont$\downarrow$ \\
\midrule
all & conv\_term & \textbf{0.07} & 0.46 & 0.09 & 0.43\\
 & multi\_tag & 0.39 & 0.81 & 0.32 & 0.82\\
 & multi\_term & 0.34 & 0.48 & 0.36 & 0.43\\
 & multi\_triple & 0.10 & 0.54 & 0.08 & 0.51\\
 & single\_code & 0.07 & \textbf{0.43} & \underline{\textbf{0.06}} & \underline{\textbf{0.39}}\\
 & single\_tag & 0.40 & 0.81 & 0.42 & 0.81\\
 & single\_term & 0.08 & 0.52 & 0.09 & 0.44\\
\midrule
7best & conv\_term & 0.08 & 0.52 & 0.10 & 0.46\\
 & multi\_tag & 0.39 & 0.82 & 0.49 & 0.81\\
 & multi\_term & 0.27 & 0.51 & 0.32 & 0.45\\
 & multi\_triple & 0.08 & 0.53 & 0.09 & 0.49\\
 & single\_code & \textbf{0.07} & \textbf{0.46} & \textbf{0.06} & \textbf{0.43}\\
 & single\_tag & 0.40 & 0.81 & 0.47 & 0.81\\
 & single\_term & 0.08 & 0.54 & 0.10 & 0.45\\
\midrule
term & multi\_term & 0.29 & 0.50 & 0.32 & 0.45\\
 & single\_term & \textbf{0.09} & \textbf{0.47} & \textbf{0.10} & \textbf{0.43}\\
\midrule
only & conv\_term & 0.08 & 0.52 & 0.09 & 0.48\\
 & multi\_term & 0.26 & 0.50 & 0.28 & 0.49\\
 & multi\_triple & \underline{\textbf{0.04}} & \underline{\textbf{0.38}} & 0.10 & \textbf{0.46}\\
 & single\_term & 0.11 & 0.54 & \textbf{0.09} & 0.55\\
\bottomrule
\end{tabular}
\caption{Difference of performances for nested/discontinuous and normal entities for each base model, training configuration, and format. Lowest difference for each model is in \textbf{bold}. \underline{Underlined} values are lowest difference overall.}
\label{tab:complex_results}
\end{center}
\end{table*}

Even though complex entities represent only a small proportion of the entities, they are important in the biomedical domain. 
All the previous studies on generative NER do not take into account complex entities (i.e., nested or discontinuous). In Table \ref{tab:complex_results}, it can be seen that overall nested entities are correctly recognized even if there is still a gap with not-complex entities. However, this is not the case for discontinuous entities. It seems that no particular format  handles the complex entities better than others. The large gap between the performance of \emph{multi\_span} and \emph{single\_span} could be explained because we exclude these formats for PGxCorpus what restrict the experiment to  simpler nested entities to recognize.

\subsection{Qualitative Analysis of span errors}
\label{sec:span_errors}

We investigated why span experiments was not working properly. To this end, we examined several erroneous outputs in detail. In addition to the usual issues, such as unrecognized and spurious entities, we found that many errors arise from shifts in the span boundaries. Examples \ref{"span_examples"} illustrate this phenomenon. Although the observed shifts seem to indicate that the shift is always negative, no constant value emerges, and generally the shift ranges between 1 and 5.

These observations suggest that smaller models may lack the capacity to reliably manipulate character-level spans. Future work involving larger models, as well as simpler span manipulation tasks, could enable a more precise assessment of the necessary model capacity for accurate character-span handling.

\begin{lstlisting}[breaklines=true, breakindent=0pt, basicstyle=\ttfamily, showstringspaces=false, caption="Span shift examples", label="span_examples"]

Text: In comparison , by day 24 , the majority of groups also treated with prednisolone displayed significantly less corneal clouding and neovascularization .
References: corneal (111:118)
Candidates: less co (106:113)
Shift: -5

Text: A good side effect - I take the lipitor at night because my husband had heard it works better taken at night .
References: lipitor (32:39)
Candidates: e lipit (30:37)
Shift: -2

Text: In one group of hemiparkinsonian rats ( n = 5 ) , caffeine caused a dose - dependent recovery of the contralateral forepaw stepping : ED50 = 2 . 4 mu mol / kg / day [ 95 % CI , 1 . 9 - 3 . 1 ] ) , reaching its maximum at the dose of 5 . 15 mu mol / kg / day .
References: feine ca (50:58)
Candidates: caffeine (53:61)
Shift: -3

\end{lstlisting}

\section{Conclusion}
\label{sec:conclusion}

This study demonstrates that instruction-tuned, lightweight LLMs can achieve competitive performance in biomedical G-NER, challenging the dominance of larger-scale models. We identified the most effective output formats (formats conv\_term and multi\_triple) for representing biomedical entities, including complex cases such as nested and discontinuous entities. Furthermore, we showed that training the model on multiple formats simultaneously does not degrade performance, offering increased flexibility and, in some instances, improving the model’s predictions. These findings suggest promising avenues for future work, including extending this approach to other IE tasks, such as relation extraction, and exploring whether multi-task formats, such as triples, may be particularly effective in some settings.


\clearpage
\newpage
\clearpage

\section*{Limitations}
\label{sec:limitations}

We focused only on two models, although many other exist. For instance, investigating BioMistral-7B, which is pre-trained on medical data would have been interesting; however, to our knowledge, no lightweight version of this type of LLM is currently available. Additionally, we did not train with the least performing formats, opting instead to concentrate on the most promising ones to limit computational costs. We also did not attempt to optimize the training hyper-parameters, instead using those from previous work that we deemed appropriate. 
Finally, we chose models that had already been instruction-tuned, although it would have been equally valuable to assess the performance of basic pre-trained models. These aspects of our study represent potential directions for future work.

\section*{Acknowledgments}

This work benefits from a PhD funding from AI4IDF. This work was granted access to the HPC resources of IDRIS under the allocation AD011016312 made by GENCI.

\section*{Bibliographical References}\label{sec:reference}

\bibliographystyle{lrec2026-natbib}
\bibliography{biblio}


\appendix
\label{sec:appendix}
\section{Appendix}

\subsection{Training hyperparameters}

Table \ref{tab:hyperparameters} lists the most important hyperparameters used for training. These were chosen to align with previous works, as explained above. The complete list of hyperparameters can be found on the GitHub repository: 

\begin{table}[!h]
    \begin{center}

\begin{tabular}{@{}lr@{}}
\toprule
\multicolumn{2}{c}{Training hyperparameters} \\ \midrule
\#epoch & 15 \\
\#GPU & 4 \\
train batch size & 2 \\
gradient accumulation & 8 \\
learning rate & 1.00E-05 \\
optimizer & \multicolumn{1}{l}{adamw} \\
learning rate schelduder & \multicolumn{1}{l}{constant} \\
warmup & \multicolumn{1}{l}{no} \\
max new token & 128 \\
eval interval & \multicolumn{1}{l}{epoch} \\
eval batch size & 8 \\
eval metric & \multicolumn{1}{l}{micro-f1} \\ \bottomrule
\end{tabular}
\caption{Most important training hyperparameters}
\label{tab:hyperparameters}
    \end{center}

\end{table}

\subsection{Energy consumption}

All experiments were conducted using four A100s (80 GB). Table \ref{tab:consumption} lists the number of training hours for each experiment, and an estimate of power consumption. This estimate is calculated using the Green Algorithms calculator\footnote{\href{https://calculator.green-algorithms.org/}{https://calculator.green-algorithms.org/}}, based on the GPU configuration used and the location (Europe, France).

    \begin{table*}[!h]
    \begin{center}
    \begin{tabular}{@{}lrrrrrr@{}}
    \toprule
     & \multicolumn{3}{l}{Qwen2.5-0.5B} & \multicolumn{3}{l}{Llama-3.2-1B} \\ \midrule
     & \multicolumn{1}{l}{HH:MM:SS} & \multicolumn{1}{l}{gCO2e} & \multicolumn{1}{l}{kWh} & \multicolumn{1}{l}{HH:MM:SS} & \multicolumn{1}{l}{gCO2e} & \multicolumn{1}{l}{kWh} \\ \midrule
    all & 02:23:46 & 234.47 & 4.57 & 03:57:50 & 388.6 & 7.58 \\
    7best & 02:11:45 & 214.8 & 4.19 & 03:47:43 & 372.2 & 7.26 \\
    conv\_term & 01:34:26 & 154.13 & 3.01 & 02:36:45 & 255.79 & 4.99 \\
    multi\_triple & 02:07:55 & 208.24 & 4.06 & 03:44:02 & 367.29 & 7.16 \\
    single\_term & 01:43:38 & 168.89 & 3.29 & 02:51:37 & 280.38 & 5.47 \\
    multi\_term & 02:03:28 & 201.68 & 3.93 & 03:31:07 & 345.97 & 6.75 \\
    term\_ner & 01:55:18 & 188.56 & 3.68 & 03:02:57 & 298.42 & 5.82 \\ \bottomrule
    \end{tabular}
    \end{center}
    \caption{Estimation of the energy consumption of each experiments}
    \label{tab:consumption}
    \end{table*}

\end{document}